\documentclass[preprint]{vgtc}               

\ifpdf
  \pdfoutput=1\relax                   
  \usepackage{graphicx}                
  \usepackage{subcaption}  
  \DeclareGraphicsExtensions{.pdf,.png,.jpg,.jpeg} 
\else
  \usepackage{graphicx}                
  \DeclareGraphicsExtensions{.eps}     
\fi%

\graphicspath{{figures/}{pictures/}{images/}{./}} 

\usepackage{microtype}                 
\PassOptionsToPackage{warn}{textcomp}  
\usepackage{textcomp}                  
\usepackage{mathptmx}                  
\usepackage{amsmath}
\usepackage{times}                     
\usepackage{cite}                      
\usepackage{tabu}                      
\usepackage{booktabs}                  
\usepackage{placeins}
\usepackage{stfloats}  
\usepackage{url}
\usepackage{multicol} 

\onlineid{0}

\vgtccategory{Research}

\usepackage{color}

\preprinttext{This is a preprint version of the manuscript accepted to ENPT XR at IEEE VR 2025.}

\title{Mesh2SLAM in VR: A Fast Geometry-Based SLAM Framework for Rapid Prototyping in Virtual Reality Applications}

\author{Carlos A. Pinheiro de Sousa\thanks{e-mail: carlos.pinheiro-de-sousa@uni-konstanz.de}\\ %
        \scriptsize University of Konstanz %
\and Heiko Hamann\thanks{e-mail:heiko.hamann@uni-konstanz.de}\\ %
     \scriptsize University of Konstanz %
\and Oliver Deussen\thanks{e-mail:oliver.deussen@uni-konstanz.de}\\ %
     \parbox{1.4in}{\scriptsize \centering University of Konstanz }}

\abstract{

SLAM is a foundational technique with broad applications in robotics and AR/VR. SLAM simulations evaluate new concepts, but testing on resource-constrained devices, such as VR \textit{HMDs}, faces challenges: high computational cost and restricted sensor data access. This work proposes a sparse framework using mesh geometry projections as features, which improves efficiency and circumvents direct sensor data access, advancing SLAM research as we demonstrate in VR and through numerical evaluation.%

} 

\begin{document}


\firstsection{Introduction}

\maketitle

Simultaneous Localization and Mapping (SLAM) is a foundational method with broad applications in robotics and computer vision, critical for the operation of augmented and virtual reality (AR/VR) devices. SLAM algorithms rely on various sensors, such as LIDAR, cameras, and depth sensors, to build a map of the environment while simultaneously estimating the device's location. Visual SLAM (V-SLAM)\footnote{This work focuses on V-SLAM specifically, for simplicity, it will be referred to as SLAM throughout this document.} primarily uses images as input data. However, additional processing of the image stream is required to detect and match salient features between frames\footnote{Considering indirect, also known as feature-based V-SLAM methods.}. Accurate detection and matching of these features are vital for estimating camera poses between consecutive frames, known as \textit{localization}.

Simulation plays a key role in efficiently testing SLAM concepts before real-world deployment \cite{Koenig2004Gazebo}. It enables rapid prototyping with reduced noise, increased repeatability, and adjustable parameters for modeling environments, processes, and sensors. In SLAM simulation, as in real-world SLAM, image features are extracted from the generated images of the virtual environment. However, regardless of  using modern AI methods \cite{revaud2019r2d2repeatablereliabledetector,dusmanu2019d2nettrainablecnnjoint,detone2018superpointselfsupervisedpointdetection,potje2024xfeatacceleratedfeatureslightweight} or classic computer vision techniques \cite{lowe2004distinctive,inproceedings}, feature-based methods introduce not only computational overhead \cite{Muzzini2023GPU,Dhakal2022SLAMShare} but also inherent noise and ambiguities in feature matching \cite{Mikolajczyk2005Evaluation,Mikolajczyk2005Affine}.

Another major challenge in SLAM research with direct use of Head-Mounted Displays (HMDs) is the restricted access to raw sensor data on commercial AR/VR hardware. While these devices often perform SLAM for real-time \textit{localization}, this functionality is typically system-level and unavailable to users. Consequently, prototyping SLAM applications on HMDs remains limited and proprietary. Most manufacturers\footnote{Exceptions include Microsoft Hololens2 and Varjo.} restrict access to raw camera inputs, confining research to simulated sensors and environments.

To address these limitations, we propose an efficient and portable framework that performs monocular SLAM directly from \textit{runtime} virtual environments, which circumvents the need for direct sensor access. Additionally, as an alternative to image-based features for SLAM, our method performs projections of mesh geometry components; we refer to it as \textit{vertex features}. Our framework prioritizes efficiency over realism by leveraging the structural elements inherent in computer-generated environments, specifically \textit{polygonal meshes}.

To our knowledge, this is the first real-time SLAM system to utilize polygonal mesh vertices as features in a virtual reality context that runs directly on HMD devices. It has three advantages:
\begin{itemize}
\item 
First, our approach eliminates feature association errors, enhances position estimation accuracy and boosts runtime efficiency, making it also a potential candidate for use as ground-truth in related applications. 

\item Second, this SLAM framework is capable of running \textit{standalone} (without a tethered connection to a personal computer for heavy-lifting), directly as a user application on low-budget off-the-shelf HMDs, bypassing the need for real sensor input.

\item Third, our framework allows the use of arbitrary 3D meshes without the need of textures.
\end{itemize}

As a result, our work expands research opportunities and prototyping for SLAM beyond VR, encompassing robotics and broader computer vision applications. It is particularly suited for SLAM research and prototyping in virtual simulation environments \cite{dosovitskiy2017carlaopenurbandriving,Koenig2004Gazebo}.

 After detailing the system and performance metrics, we showcase its application in a single-user VR interaction, mapping a one-to-one sparse representation of the virtual environment through user motion. 

\section{Related Work}

The use of image features for visual odometry dates back to NASA’s 1980s Mars exploration programs \cite{moravec_obstacle_1980}. Feature-based real-time visual odometry and reconstruction, or \textit{localization and mapping}, became practical decades later with PTAM \cite{klein_parallel_2007}, which reduced computational costs by employing parallel threads for tracking and mapping.

Feature-based methods, often called indirect methods, rely on a front-end module to extract features from image sequences. While robust to photometric changes and large baselines, they demand high computational resources, challenging real-time performance requirements.

Prominent SLAM frameworks, such as ORB-SLAM [18], aim to mitigate these ongoing challenges by leveraging efficient feature extraction methods like ORB [22]. ORB-SLAM’s modular design has made it widely adopted, enabling extensions and fostering innovation in the SLAM field.

Research on SLAM specifically for virtual reality (VR) remains limited. Many existing approaches rely on simulation-based methods that either do not capture the user’s perspective for mapping or simply use a device’s built-in SLAM framework for localization. For instance, \cite{bettens2020unrealnavigation} introduced a SLAM testing platform in virtual environments—described as VR—yet did not employ physical head-mounted devices (HMDs). Similarly, \cite{zins2022oaslamleveragingobjectscamera} used SLAM on the HoloLens 2, but it was driven by the system’s built-in SLAM, illustrating an application of SLAM rather than advancing new SLAM techniques. In another study, \cite{10.1007/s00371-022-02530-1} developed visual–inertial SLAM for hand controllers rather than from the rendered VR environment itself.

In contrast, our proposed Mesh2SLAM performs real-time SLAM by directly leveraging polygonal meshes from virtual environments and can operate entirely in VR, with the user as the primary mapping agent.

\section{Method}
\subsection{Overview}

We developed Mesh2SLAM drawing inspiration from the simplicity and dual parallel modules of PTAM (Parallel Tracking and Mapping). Additionally, we adapt the initialization, tracking, and mapping processes from ORB-SLAM2, with such modules tailored to integrate with our own feature processing method.

Given the growing importance of resource-constrained devices in SLAM research \cite{Cadena16tro-SLAMfuture}, our method is designed to be lightweight, portable and more independent of third-party libraries. Core component modules that process image features are replaced with our \textit{vertex feature} processing approach, furthermore, particularly for visualization, it includes its own, optional, OpenGL ES visualization engine allowing greater portability. 

\begin{figure}[ht]
    \centering
    \includegraphics[width=0.4\textwidth]{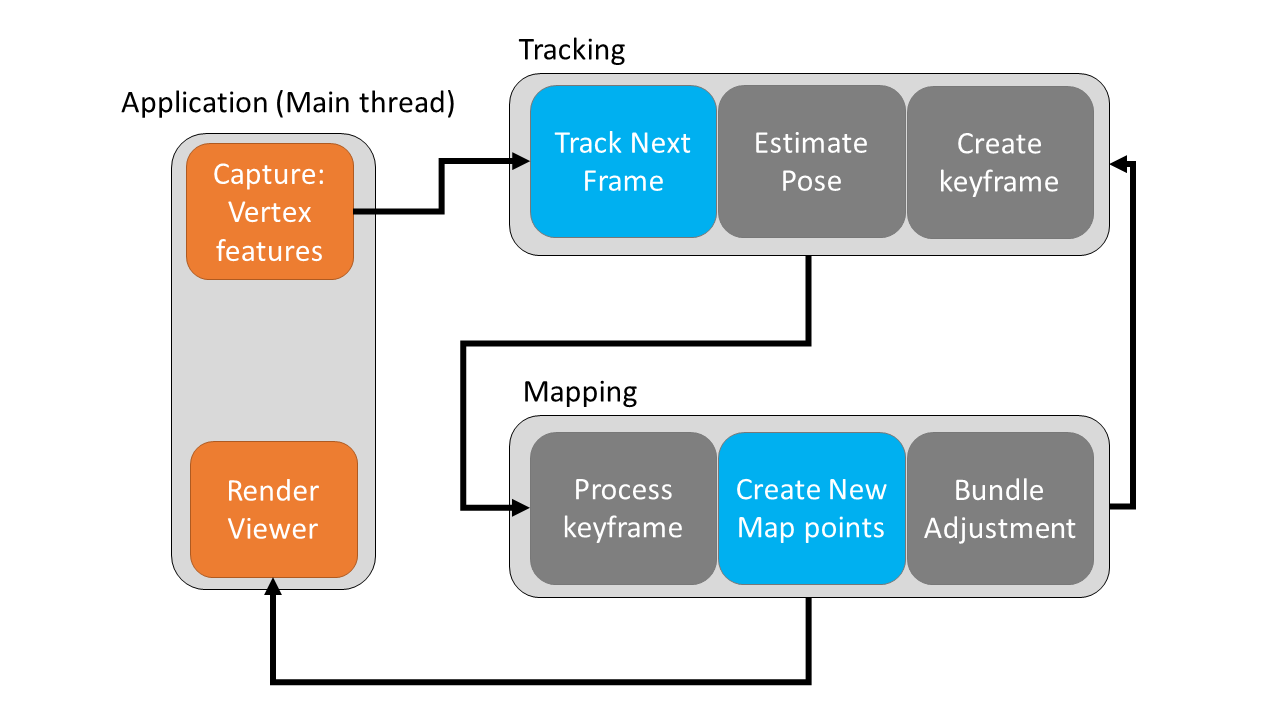}
    \caption{System overview, showing main threads; highlighted (blue) are related to vertex features processing, highlighted (orange) are executed in application main thread, arrows represent event triggers}
    \label{fig:system1}
\end{figure}

As shown in Figure \ref{fig:system1}, Mesh2SLAM operates concurrently across the main application thread and two dedicated SLAM threads: \textit{Tracking} and \textit{Mapping}. Frames are captured from the scene geometry as vertex features and processed by the front-end \textit{Tracking} module. The back-end \textit{Mapping} module then performs posterior optimization, including windowed bundle adjustment \cite{6096039}, to minimize errors and achieve the best fit for frame poses and mapped points. 

The multi-threaded design highlights the importance of maintaining high performance and independence from the main rendering thread in our system.

\begin{figure}[ht]
    \centering
    \includegraphics[width=0.4\textwidth]{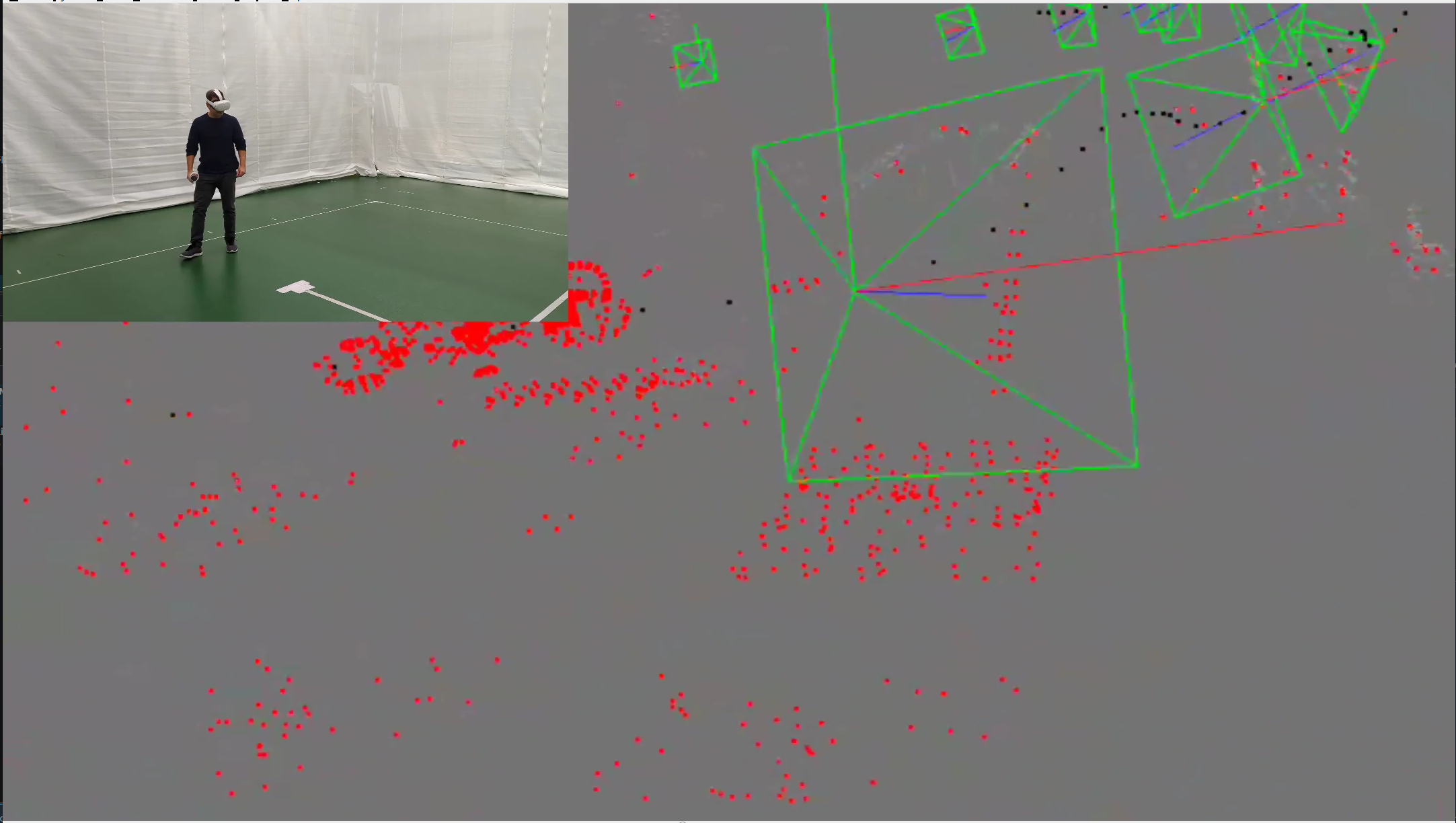}
    \caption{A screenshot live VR while running Mesh2SLAM.}
    \label{fig:VRLive}
\end{figure}

Finally, the generated map is rendered back on the main thread, with frame poses and the point cloud map displayed in a first-person perspective, as shown in Figure \ref{fig:VRLive}.

 The following sections detail the key processes specific to our method.
\subsection{Frame Capture}
 For the typical operation, the mesh model is loaded into memory at start up. The sequence of vertex components is registered ensuring uniquely identified vertices by stored index.

\begin{figure}[ht]
    \centering
    \includegraphics[width=0.3\textwidth]{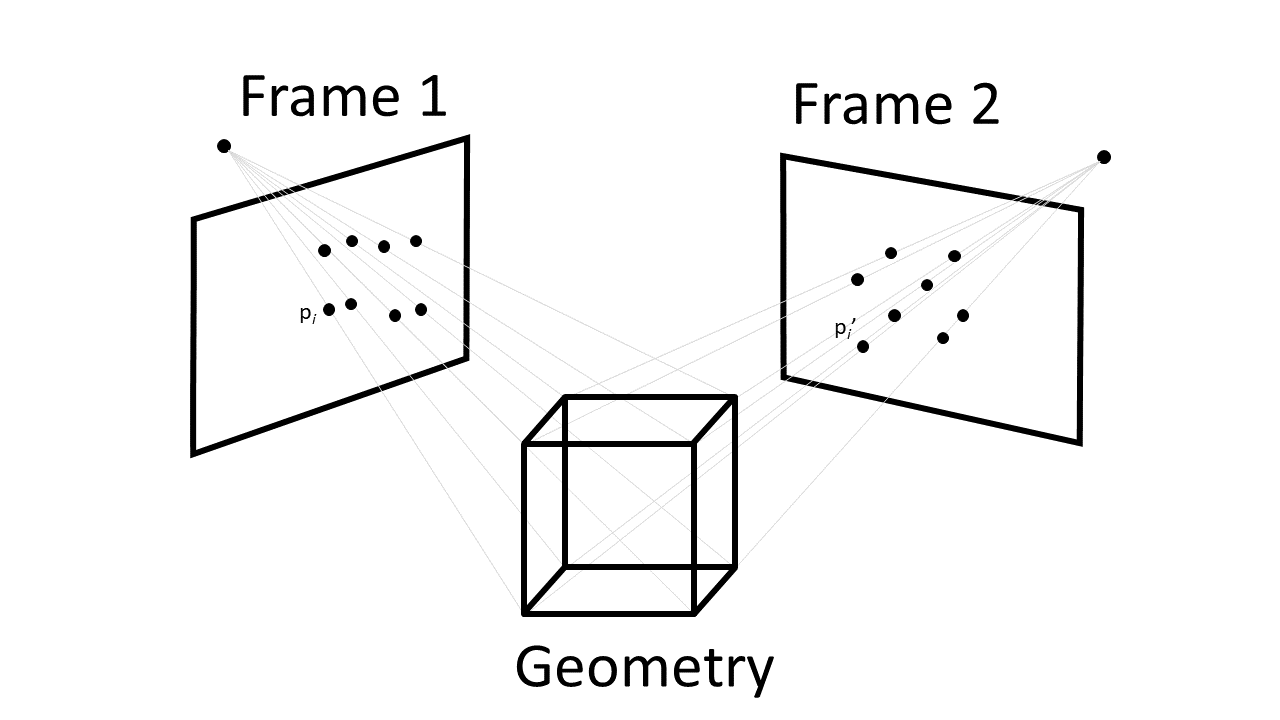}
    \caption{Direct geometric projection of vertices to the camera view in Mesh2SLAM.} 
    \label{fig:geometryFigure}
\end{figure}

At each iteration of the main rendering loop, the model's vertices are rendered or \textit{captured} as vertex features and are input into the SLAM system. The capture process utilizes a custom  \textit{compute shader} \cite{opengl_compute_shader} to project vertices into the virtual camera's image plane for each frame see Fig. \ref{fig:geometryFigure}.
 
The resulting vertex features are output as a list of features with their specific identifiers and image coordinates. The details of the computation are clarified:

Let $\mathbf{p}_i$ represent the position of the $i$-th vertex in world space:

\[
\mathbf{p}_i = \begin{pmatrix} x_i \\ y_i \\ z_i \end{pmatrix}
\]

 The model vertices are projected into image space for every frame following traditional model-view-projection matrix operations \eqref{eq:vclip}, where, \textit{P} is the perspective projection, \textit{V} is the camera view matrix and \textit{M} is the model matrix and $\mathbf{p}_i$ now represented in \textit{homogenous coordinates} 
\vspace{-2em}
\begin{multicols}{2}
\begin{equation}
\mathbf{v}_{\text{clip}} = \mathbf{P} \cdot \mathbf{V} \cdot \mathbf{M} \cdot 
\begin{pmatrix} x_i \\ y_i \\ z_i \\ 1 \end{pmatrix}
\label{eq:vclip}
\end{equation}
\break
\begin{equation}
\mathbf{v}_{\text{view}} = \mathbf{V} \cdot \mathbf{M} \cdot 
\begin{pmatrix} x_i \\ y_i \\ z_i \\ 1 \end{pmatrix}
\label{eq:vview}
\end{equation}
\end{multicols}

 Complementarily,  \textit{z} components, representing depth with respect to the camera, are kept in view space \eqref{eq:vview} in order to assert depth thresholds; Depth can be used to include only vertexes that are within acceptable range.

The values for clip and view can now be used to compute the normalized screen coordinates (ndc) \eqref{eq:ndc_x},\eqref{eq:ndc_y} and depth \eqref{eq:ndc_z}:
\begin{subequations}
\label{eq:ndc}
\begin{align}
x_{\text{ndc}} &= \mathbf{v}_{\text{clip}}.x \cdot \mathbf{v}_{\text{clip}}.w^{-1}, \label{eq:ndc_x} \\
y_{\text{ndc}} &= \mathbf{v}_{\text{clip}}.y \cdot \mathbf{v}_{\text{clip}}.w^{-1}, \label{eq:ndc_y} \\
z &= -\mathbf{v}_{\text{view}}.z. \label{eq:ndc_z}
\end{align}
\end{subequations}
And finally screen coordinates \textit{u}, \textit{v} are obtained with \textit{w} and \textit{h} as width and height of screen resolution:

\begin{equation}
\begin{aligned}
u &= \left( x_{\text{ndc}} \times 0.5 + 0.5 \right) \times w, \quad
v = \left( -y_{\text{ndc}} \times 0.5 + 0.5 \right) \times h.
\end{aligned}
\label{eq:uv}
\end{equation}

\noindent As a result, a \textit{vertex feature} aggregates the screen coordinates along with the descriptor, the vertex’s unique identifier \textit{i}:
\[
\mathbf{v}_i = \begin{pmatrix} u_i \\ v_i \\ i \end{pmatrix}
\]
Next we clarify how our approach provides advantages for feature processing.

\subsection{Feature Proccesing with vertex features}
 
In feature-based SLAM processing, the association and matching of feature correspondences between frame pairs are computed by calculating the \textit{similarity} between the feature descriptors. Our approach streamlines this process by leveraging unique IDs as descriptors, i.e, a simple integer comparison, allowing for more efficient and errorless feature association. 

Traditional image-feature methods, suffer from degraded feature matching under large parallax or substantial baselines between camera poses. This is often mitigated by limiting the average normal divergence angle between frames that observe the same feature \cite{murTRO2015}. Our approach eliminates this constraint since regardless of angle, \textit{vertex features} are viewpoint-invariant. Similarly, traditional methods require scale matching to ensure consistent feature measurement across frames. With our approach, this is unnecessary, as scale invariance is inherently enforced by depth threshold limits.

\begin{figure}[ht]
    \centering
    \includegraphics[width=0.4\textwidth]{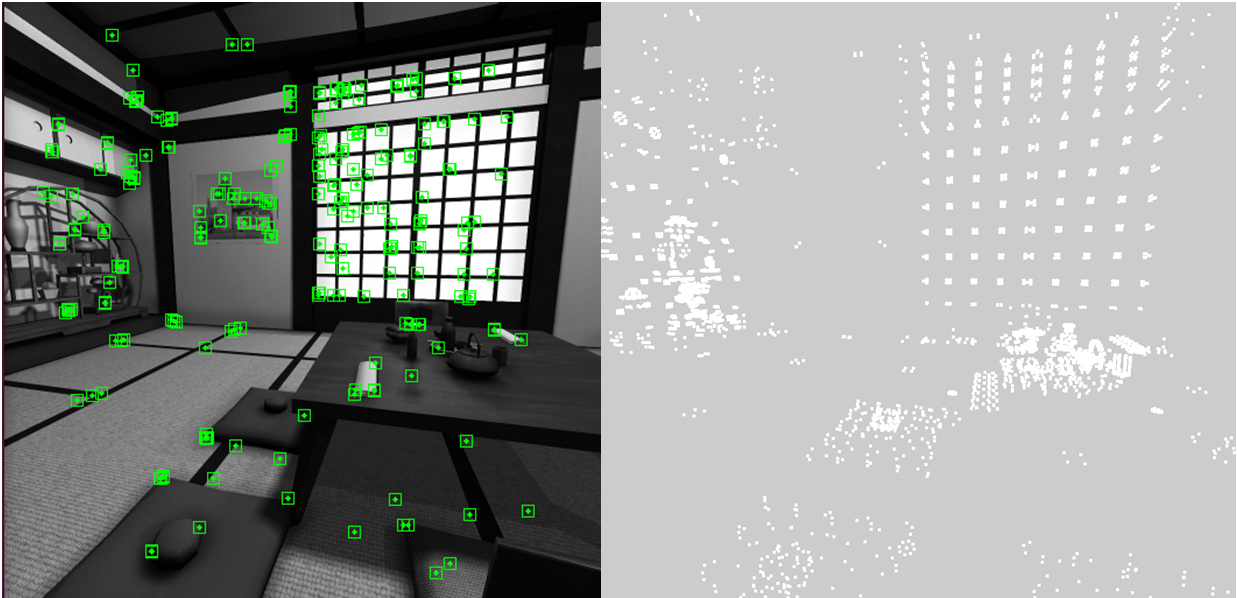}
    \caption{Left: A display of image-based features. Right: Our approach vertex features.}
    \label{fig:FeaturesORBM2S}
\end{figure}

 Figure \ref{fig:FeaturesORBM2S} illustrates a comparison between traditional image features and our approach \textit{vertex features}

 \subsection{SLAM Loop}
The system functionality follows that of ORB-SLAM2 and can be broken down to: initialization then parallel tracking and mapping. At initialization, the system attempts to create an initial map. Then once a map is initialized, Tracking and Mapping process the sequential frames containing \textit{vertex features} data in parallel. For tracking, camera pose estimations are computed with association of features with previous frame. For mapping, new map points are created by triangulation, provided that, among other conditions, enough unmapped features of previous frame match with the new frame's features observations.     

 \subsection{Viewer}
The visualization component is managed by our custom \textit{Viewer} module, which runs on the main application thread. It is updated in response to new map update events. The \textit{Viewer} rendering itself follows the frame rate of the main application, independently of SLAM performance. The result is a generated sparse point cloud map that accurately overlays the original mesh geometry as the user scans the virtual environment as seen in \ref{fig:FeaturesORBM2S}.

For this application, \textit{Viewer} has been integrated with \textit{OpenXR} and will be described next.

 \subsection{OpenXR Integration}

\begin{figure}[ht]
    \centering
    \includegraphics[width=0.1\textwidth]{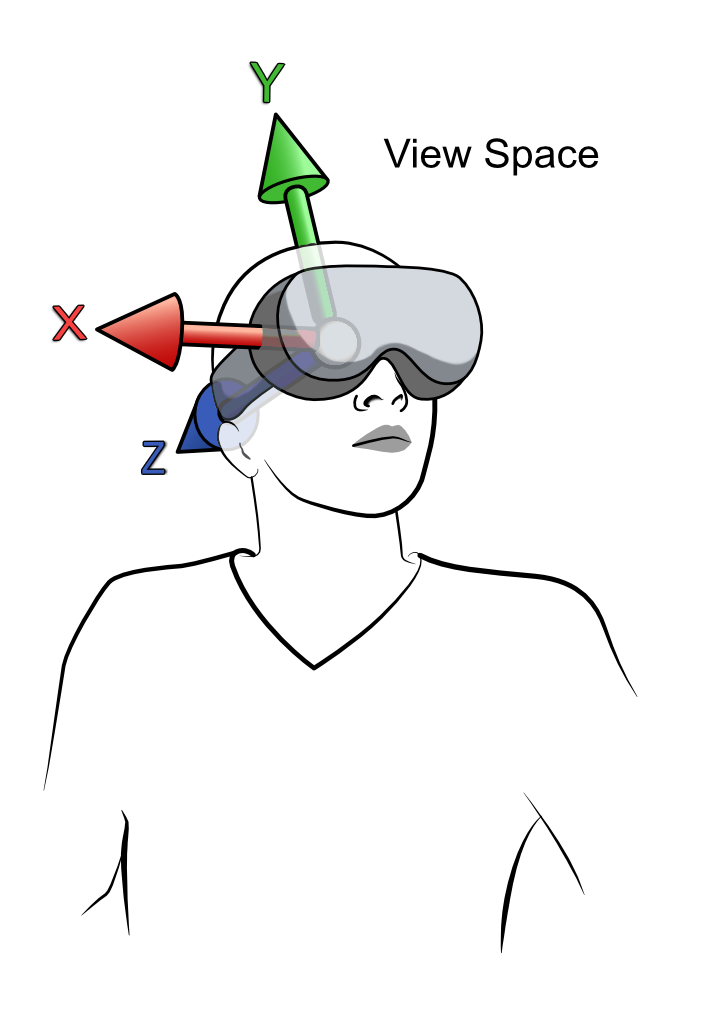}
    \caption{\textit{View Space}, used as reference for vertex feature extraction (obtained from \cite{Khronos:2019}).}
    \label{fig:mesh1}
\end{figure}

For proper application in virtual reality, the camera pose or reference used for geometry projection should match that of the user's perspective or viewpoint. In this monocular setup, the reference is centered in the device's View Space, (Fig. \ref{fig:mesh1}) which serves as a frame of reference as defined by the OpenXR API.

\section{Experiments and Results}
\label{sec:experiments}
In this section, we demonstrate the efficiency and effectiveness of our method under several important metrics: 1) Efficiency of Vertex Feature Extraction, 2) System Performance and 3) Virtual Reality Practical Evaluation

\subsection{Efficiency of Vertex Feature Extraction}

We conducted experiments using polygonal meshes ranging from 500 to 2 million vertices on both a mid-range notebook (Intel(R) Core(TM) i7-1065G7 CPU @ 1.30GHz, Intel® Iris Plus Graphics G7, 16 GB of RAM) and a low-budget (mid-range) HMD, the Oculus Quest 2 \cite{MetaUnityPerf}. 

The performance metric used is the average time in \textit{milliseconds} from the dispatch of the compute shader program to the output of the compute shader of resulting vertex feature components into vector storage, additionally, since this process runs on the main application thread, i.e. rendering loop, the nominal application framerate is also measured in frames-per-second (FPS). The results are presented in Table \ref{tab:performance_table} below:

\begin{table}[ht]
    \centering
    \begin{tabular}{|c|c|c|}
        \hline
        \textbf{Vertex Count} & \textbf{PC (ms)/ FPS} & \textbf{Quest 2 (ms) / FPS} \\ \hline
        600         & 12 /75   & 0.006 /72   \\ \hline
        60 000         & 12 /75   & 2  /72 \\ \hline
        240 000        & 12 /75   &  8 /72   \\ \hline
        480 000        & 12 /75   & 14 /58  \\ \hline
        2 000 000   & 12 /75  & 60 /14 \\ \hline
    \end{tabular}
    \caption{Median time (in ms) for 1000 runs for different counts of mesh vertex feature computation on a personal computer (PC) and Quest 2.}
    \label{tab:performance_table}
\end{table}

While the personal computer maintained stable performance across all vertex counts, the Meta Quest 2 exhibited performance degradation beyond 400,000 vertices, as shown by the drop in frame rate. We believe this correlates with Quest's performance budget highlighted in \cite{MetaUnityPerf}, also because no frustum culling is performed on our side for these tests. Nonetheless, the performance of vertex feature extraction method far surpasses that of traditional image-based feature extraction, which typically does not exceed a few thousand features extracted per image. 

\subsection{System Performance}
\subsubsection{Experiment Setup}

\textbf{Virtual environment}: 
Next we present an evaluation of localization accuracry of Mesh2SLAM in comparison to the baseline ORB-SLAM2. The overall tracking behaviour with varying camera image input frame-rates, 15, 30, 60 and 75 and accuracy are tested on a PC running computer graphics-generated environment specifically designed for both methods.

\begin{figure}[ht]
    \centering
    \includegraphics[width=0.3\textwidth]{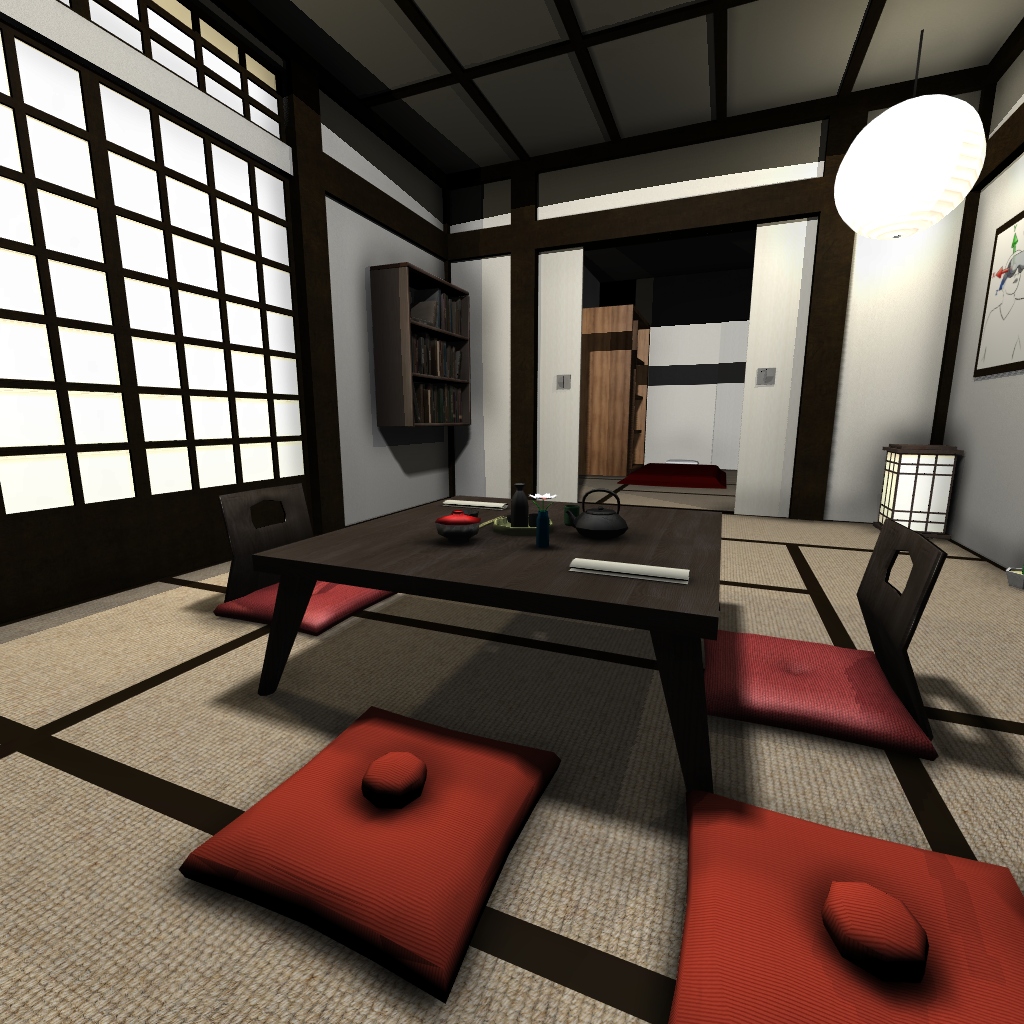}
    \caption{Pre-rendered scene, used for image feature extraction.}
    \label{fig:tatamiRender}
\end{figure}

For image feature-based method, the scene is pre-rendered as an image sequence at resolution of 1024 x 1024 pixels and additional lighting and textures are used aiming for a realistic look (see Fig. \ref{fig:tatamiRender}). Concurrently, the environment consists of a scene composed by polygonal meshes used as source for the vertex feature extraction for our method.  

\begin{figure}[htbp!]
    \centering
    \includegraphics[width=0.7\linewidth]{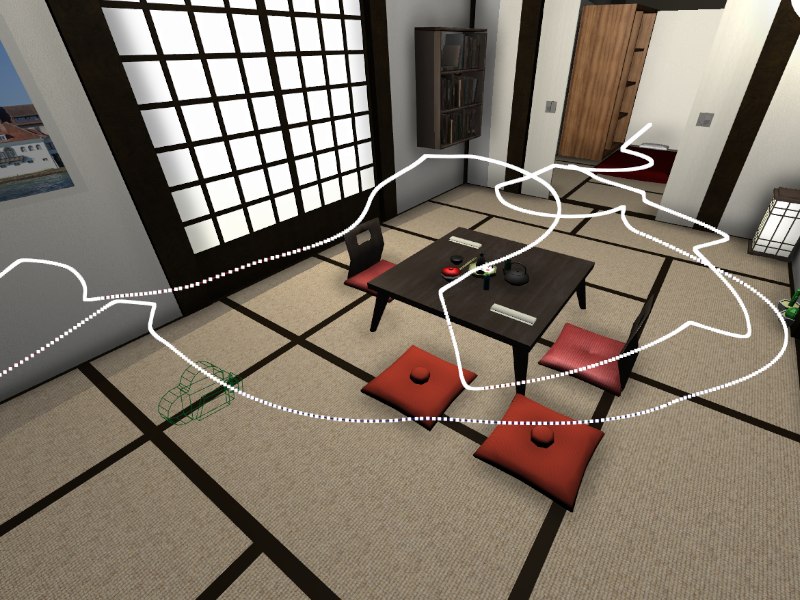}
    \caption{Prerecorded camera trajectory/poses which is re-used for evaluation.}
    \label{fig:cameraTrajectory}
\end{figure}

\textbf{Camera setup: } 
For the evaluation, a first-person style camera motion is recorded and is re-used (show in figure \ref{fig:cameraTrajectory}). The camera projection uses configuration with a field-of-view of 90 degrees and no lens-distortion.

\vspace{-1em} 

\textbf{ORB-SLAM2 Performance: } 
ORB-SLAM2 has been altered for asynchronous input frame-rate i.e. with possible frame skipping if processing time takes longer than expected for next sequential frame. Such adaptation assures a fair comparison with our method which processes input frames in asynchronous manner as well.

With the original source image resolution of 1024 by 1024 pixels, ORB-SLAM2 could only operate at an average of 15 FPS with eventual loss of tracking. At higher frame-rates, 30, 60 and 75 was only possible by halving of image resolution and led to recurring loss of tracking with abrupt movements. Further reduction in input image resolution deemed unfeasible due to poor system performance.

\textbf{Mesh2SLAM Performance: }
Our method functions effectively across all tested frame rates: 30, 60, and 75 FPS (The test was conducted on a PC where the refresh rate was capped at 75 Hz due to hardware limitations). 

In table \ref{tab:ATE ORBSLAM Mesh2SLAM} we briefly present the results which reflect the degradation of localization with increasing frame-rates. The evaluation of SLAM performance is typically measured by comparing estimated camera pose trajectories to ground truth \cite{8593941,Kmmerle2009OnMT}. These values present the Absolute Trajectory Error, RMSE. As expected, Mesh2SLAM has very high accuracy result, standing around an order of magnitude lower error than the baseline method for higher frame rate. A visual inspection of the tracking quality through the trajectory of camera sequence is presented next, with figure in Appendix ~\ref{sec:appendix_figs}.   

\begin{table}[ht]
    \centering
    \begin{tabular}{|c|c|c|}
        \hline
        \textbf{Frame Rate} & \textbf{Mesh2SLAM (RMSE)} & \textbf{ORB-SLAM2 (RMSE)} \\ \hline
        30  & 0.037 $\pm$ 0.026  & 0.124 $\pm$ 0.065\\ \hline
        60  & 0.065 $\pm$ 0.039  & 0.147 $\pm$ 0.060\\ \hline
        75  & 0.095 $\pm$ 0.060  & 2.011 $\pm$ 1.033\\ \hline
    \end{tabular}
    \caption{Comparison of performance for systems Mesh2SLAM and ORB-SLAM2 at different frame rates with associated errors.}
    \label{tab:ATE ORBSLAM Mesh2SLAM}
\end{table}

\subsubsection{Absolute Trajectory Error Visual Inspection}

As shown in Figure~\ref{fig:ATE_M2S_ORB} (see Appendix~\ref{sec:appendix_figs}) and summarized in the table above, the localization accuracy of our method significantly outperforms ORB-SLAM2 when compared to the ground truth.  We include frame rates of 30~FPS to match ORB-SLAM2’s ideal rate and 75~FPS as typically desired for VR. Notably, at the end of the sequence at 75~FPS, ORB-SLAM2 loses tracking entirely, indicated by missing aligned red line segments in the figure.

\subsection{Virtual Reality Practical Evaluation}

For a practical functionality with a real-time interactive demonstration we evaluate the application running as \textit{standalone} on a Meta Quest 2. It starts with the user at scene origin in the computer-generated environment. As SLAM performs in real-time, the mapping provides the reconstruction of the scene as sparse set of red points, which correspond to the mapped vertexes of the virtual environment and the corresponding user poses at each instant. Subsequently as the user moves and interactively looks around, the scene is mapped accordingly. Finally, on a qualitative description, this demo runs at 72 FPS with no noticeable jitter or freezing; vertex feature extractions have no practical overhead and the SLAM threads run independently from the main application rendering thread. 

\section{Scope and Constraints}

Mesh2SLAM is limited to virtual environments, which restricts its applicability to real-world scenarios. Additionally, unbalanced spatial density or clustering of mesh vertices can impact tracking performance. Although the scene is currently treated as a single mesh, future extensions could handle large-scale environments with multiple meshes, which may be achieved through other optimization, sorting, or culling mechanisms. While our evaluation focused on the Meta Quest 2, we expect testing on other HMDs will likely yield similar results and provide a more comprehensive assessment.

\section{Conclusion}

In this work, we presented Mesh2SLAM, a fast and efficient geometry-based SLAM framework for prototyping that operates on HMDs in virtual reality. Its main novelty lies in utilizing mesh geometry components, \textit{vertex features}, and leveraging GPU acceleration to enhance position estimation accuracy and drastically reduce computational overhead. To our knowledge, it is the first real-time SLAM method running in a virtual environment as a \textit{standalone} application, without tethering, on low-budget, off-the-shelf HMDs.

As a SLAM simulation, Mesh2SLAM maintains SLAM functionality while bypassing issues in image-based methods that hinder prototyping. In the context of XR development, its lightweight sparse map representation is well-suited for localization-focused tasks, including tracking user movement, supporting spatial interactions, validating pose estimation methods, and benchmarking other localization algorithms. Additionally, future considerations include extending it to collaborative localization or multi-agent SLAM. These capabilities make it a valuable tool for efficiently testing and refining concepts, driving advancements in XR research.

\clearpage  

\appendix
\section{Additional Figures}
\label{sec:appendix_figs}

Below we present additional illustrations of the Absolute Trajectory Error (ATE) 
for ORB-SLAM2 and Mesh2SLAM at different frame rates. 
This figure highlights the trajectory alignment and differences 
between the estimated and ground-truth camera poses.

\noindent
\includegraphics[width=\textwidth]{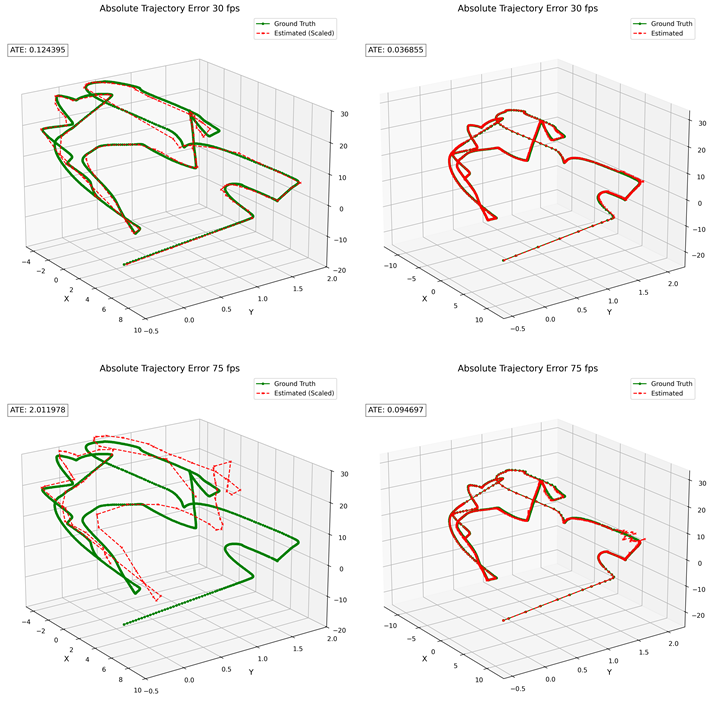}

\begin{center}
\parbox{\textwidth}{
\captionof{figure}{ATE comparison at 30 FPS and 75 FPS for ORB-SLAM2 (left) and Mesh2SLAM (right).}
\label{fig:ATE_M2S_ORB}}
\end{center}

\clearpage  
\bibliographystyle{abbrv-doi}
\bibliography{Mesh2SLAM}

\end{document}